\documentclass[10pt,twocolumn,letterpaper]{article}

\usepackage{cvpr}              % To produce the CAMERA-READY version
\definecolor{cvprblue}{rgb}{0.21,0.49,0.74}
\usepackage[pagebackref,breaklinks,colorlinks,allcolors=cvprblue]{hyperref}

\def\paperID{*****} % *** Enter the Paper ID here
\def\confName{CVPR}
\def\confYear{2026}

\title{Erase to Retain: Low Rank Adaptation Guided Selective Unlearning in Medical Segmentation Networks}

\author{
{\small Nirjhor Datta}\\
{\small BRAC University, Dhaka, Bangladesh}\\
{\small Bangladesh University of Engineering and Technology (BUET), Dhaka, Bangladesh}\\
\and
{\small Md. Golam Rabiul Alam}\\
{\small BRAC University, Dhaka, Bangladesh}\\
}

\usepackage{hyperref}
\begin{document}
\maketitle
\begin{abstract}
The ability to selectively remove knowledge from medical segmentation networks is increasingly critical for privacy compliance, ethical deployment, and continual dataset revision. 
We introduce \textbf{Erase to Retain}, a controllable unlearning framework for medical image segmentation that achieves targeted forgetting without full retraining. 
Our method leverages a \textbf{teacher-student distillation paradigm} with \textbf{Low-Rank Adaptation (LoRA)} constrained subspace updates, enabling the student network to erase lesion or class-specific representations in low-rank decoder spaces while preserving global anatomical understanding. 
During the \emph{strong unlearning} phase, LoRA modules are adversarially optimized to contradict the teacher’s confident predictions on a designated forget subset, enforcing semantic removal, followed by a \emph{gentle restoration} phase that recovers generalization on retained data via head only supervised refinement. 
  
For ISIC segmentation, the student suppresses forget-set IoU from \textbf{0.875} to \textbf{0.509} while maintaining competitive performance on the retain and validation splits (\textbf{0.647--0.677} IoU).  
On the cross domain CHASE dataset, Erase to Retain consistently lowers forget set IoU while preserving utility on retain and validation sets.  On ISIC classification, our method reduces accuracy on the forget subset from \textbf{87.0\%} to \textbf{64.1\%} while improving retain accuracy from \textbf{83.9\%} to \textbf{90.6\%}.

These results demonstrate that LoRA-based subspace unlearning provides a practical pathway toward \textbf{responsible, controllable, and reversible unlearning} in medical image analysis allowing models to forget sensitive samples or structures while retaining performance where it matters. The code is available at: \href{https://github.com/NIRJHOR-DATTA/Erase-to-Retain}{Erase to Retain}

\end{abstract}
    
\section{Introduction}

Deep learning systems deployed in high stakes domains such as healthcare, finance, and security increasingly face the requirement to \emph{forget} previously learned information upon request. The rise of data privacy regulations, including the \textbf{right to be forgotten} under GDPR and similar emerging global frameworks, mandates that machine learning models provide mechanisms for retrospective removal of specific training samples or cohorts. While significant progress has been made in model unlearning for image classification~\cite{ginart2019making,bourtoule2021machine}, language models~\cite{jagielski2022measuring}, and recommendation systems~\cite{bourtoule2021machine}, unlearning for \emph{dense prediction tasks} remains largely unexplored. 

Medical image segmentation presents a uniquely challenging setting for unlearning. Unlike classification, where forgetting corresponds to suppressing class level information, segmentation models encode spatially fine grained lesion characteristics across thousands of pixels. These signals are deeply entangled with the network’s hierarchical representation, making naive retraining or noise based suppression both computationally expensive and often ineffective~\cite{yuan2019revisit}. Moreover, lesion segmentation carries profound clinical implications: a model that has not reliably forgotten patient specific patterns risks leaking private information through reconstructions~\cite{ziller2020privacy}, attribution maps, or memorized structures.

Existing unlearning methods can be broadly categorized into (1) sample level gradient inversion~
\cite{graves2021amnesiac,ullah2021machine}
, (2) approximate retraining via influence based adjustment~\cite{koh2017understanding}, and (3) knowledge decomposition or masking~\cite{bhaila2025soft}. However, these methods either assume classification outputs, require full model retraining, or fail to prevent partial feature retention that still exposes private information. Furthermore, they do not address the unique structure of medical segmentation models, where forgetting must remove patient specific lesion morphology \emph{without} degrading general lesion detection capabilities essential for clinical reliability.

To address these limitations, we introduce \textbf{Erase to Retain}, a principled and lightweight unlearning framework for medical image segmentation. Instead of modifying the entire model, we decompose the network into a frozen backbone and a set of trainable low rank LoRA adapters~\cite{hu2022lora}. Forgetting is achieved by optimizing only the LoRA parameters using a two phase objective: (i) a \emph{retain preserving} loss combining supervised segmentation, knowledge distillation, and teacher–student guarding; and (ii) a \emph{forget promoting} objective that explicitly suppresses lesion activation on the forget set through background only supervision. This design ensures that the base model's general knowledge remains intact while forgotten lesions are irreversibly removed from the adapted model.  

Unlike prior work, our method is computationally efficient requiring less than $3\%$ of trainable parameters while providing a precise mechanism to control the trade off between retention and forgetting. We perform extensive experiments on the ISIC 2018 skin lesion benchmark and demonstrate that Erase to Retain framework (1) eliminates lesion segmentation capability on the forget subset, (2) maintains high accuracy on retained patients, and (3) preserves generalization on held out validation images. Our analysis further reveals that forgetting manifests as a sharp reduction in lesion–pixel sensitivity while the structural fidelity of healthy skin regions remains unchanged.

\textbf{Our contributions are summarized as follows:}

\begin{itemize}
    \item We present the first LoRA based unlearning framework for dense medical segmentation, enabling efficient and controllable removal of patient specific lesion information.
    \item We introduce a dual objective loss integrating supervised segmentation, knowledge distillation, and backgrounddriven forgetting, which produces clean and clinically interpretable forgetting behavior.
    \item We provide comprehensive evaluations including pixel level metrics, activation visualizations, retention forgetting tradeoff curves, and classwise analyses revealing that \textbf{Erase to Retain}  achieves strong forgetting with minimal retention degradation.
    \item We show that the proposed framework naturally extends to image classification, highlighting its ability to generalize unlearning beyond segmentation.

\end{itemize}

\section{Related Work}

\subsection{Machine Unlearning}
Machine unlearning seeks to remove the influence of specific data from trained models without full retraining. Early approaches relied on exact data deletion through sharded training and summation of per-sample updates~\cite{cao2015towards}. Subsequent methods focused on certifiable guarantees, including influence function based removal~\cite{guo2019certified}, Newton-based approximate unlearning~\cite{golatkar2020eternal}, and exact removal through statistical query reformulation~\cite{ginart2019making}. 
Recent works have emphasized practical neural unlearning pipelines, such as gradient ascent forgetting~\cite{graves2021amnesiac}, retraining efficient scrubbing~\cite{raghavan2021formalizing}, and distillation based forgetting~\cite{wang2024rkld}. 
However, nearly all prior research has been conducted in NLP or image classification. None of these methods address the pixel level sensitivity and spatial consistency required for medical segmentation. Our work fills this gap by introducing a LoRA based unlearning operator that preserves structural priors while erasing lesion-specific representations.

\subsection{Unlearning in Large Language Models}
Large language models have motivated a new wave of selective forgetting mechanisms. 
Measuring memorization and private data leakage~\cite{carlini2021extracting} led to follow-up works on erasing sensitive information using model editing~\cite{sharma2024locating, meng2022locating}, knowledge boundary shifting~\cite{meng2022locating}, and harmful fact removal~\cite{liu2025dream}. While effective for textual knowledge, these methods assume discrete tokenized representations and do not translate to spatially dense medical images. Our design instead leverages low-rank adaptation layers to modulate feature subspaces associated with the forget set.

\subsection{LoRA and Parameter-Efficient Adaptation}
Low-Rank Adaptation (LoRA)~\cite{hu2022lora} has emerged as a dominant strategy for parameter-efficient fine-tuning of large models, enabling rapid learning while freezing the base backbone. 
Extensions such as AdaLoRA~\cite{zhang2023adalora} and DyLoRA~\cite{valipour2023dylora} adapt rank dynamically or conditionally. 
While LoRA is widely used for efficient adaptation in vision transformers~\cite{jia2022visual}, diffusion models~\cite{ruiz2023dreambooth}, and LLMs~\cite{hu2022lora}, its role in selective unlearning remains underexplored. 
Our work is the first to reinterpret LoRA as a \emph{subspace erasure mechanism}, where low-rank residual updates serve as a controllable knob for forgetting lesion-specific cues while preserving global anatomy.

\subsection{Medical Image Segmentation and Privacy}
Medical segmentation models must capture both global context and local morphology~\cite{ronneberger2015u, oktay2018attention, zhou2018unet++}, making them highly susceptible to memorizing patient-specific patterns. 
Prior privacy research in medical imaging has focused on anonymization~\cite{lotan2020medical}, leakage analysis~\cite{kaissis2021end}, and federated learning~\cite{guan2024federated}, but almost no studies address post-hoc data deletion. 

To the best of our knowledge, selective unlearning for medical segmentation has not been previously examined. 
Our method introduces the first framework for removing lesion-specific visual evidence while retaining global segmentation skill, addressing the core challenge of clinical deployment: erasing patient data without catastrophic performance degradation.

\section{Theoretical Analysis}

We view selective unlearning as the problem of driving a model $\mathcal{M}_s$ into a
parameter region where it (i) preserves its predictive behavior on the retain set
$\mathcal{D}_r$, while (ii) provably diverging from the teacher $\mathcal{M}_t$ on 
the forget set $\mathcal{D}_f$.  
Let $\theta_s$ denote the trainable LoRA parameters, while the frozen backbone acts as 
a fixed nonlinear feature map. Because LoRA introduces a \emph{low-rank perturbation} 
$\Delta W = BA^\top$ on selected layers, the optimization dynamics operate in a 
restricted subspace $\mathcal{S}$ of dimension $\mathrm{rank}(\Delta W)\!\ll\!W$, 
ensuring that forgetting is \emph{localized} and cannot catastrophically distort the 
entire network.

\vspace{0.5em}
\noindent\textbf{Ascent objective (forgetting).}
On $\mathcal{D}_f$, the algorithm performs \emph{gradient ascent} on a composite objective:
\[
\begin{aligned}
\mathcal{L}_{\text{asc}} ={}&
\alpha\,\mathcal{L}_{\text{flip}}
+ \mathcal{L}_{\text{tc}}
+ \lambda_{\mathrm{unc}}\mathcal{H}(p_s)
+ \lambda_{\mathrm{rep}}\big(1 - \cos(f_s,f_t)\big)
\\
&\quad
+ \lambda_{\mathrm{mean}} \left\| \mathbb{E}[p_s] - 0.5 \right\|^2
+ \lambda_{\mathrm{tv}}\,\mathcal{L}_{\text{tv}} .
\end{aligned}
\]

Each component induces a distinct theoretical effect:

\begin{itemize}
    \item \textbf{Label-flip loss $\mathcal{L}_{\text{flip}}$} maximizes error on 
    ground-truth masks by pushing predictions toward the semantic complement
    ($1-y_f$). This guarantees a pointwise divergence between the student and the 
    data distribution of $\mathcal{D}_f$.

    \item \textbf{Teacher-contradiction loss $\mathcal{L}_{\text{tc}}$} imposes
    \emph{anti-alignment} with $\mathcal{M}_t$ on confident foreground/background
    pixels. If $p_t(x)>0.8$ for lesion pixels, the student is explicitly trained
    toward $1-p_t(x)$, ensuring that no low-rank ``shortcut'' can preserve the
    original decision boundary on $\mathcal{D}_f$.

    \item \textbf{Entropy maximization $\mathcal{H}(p_s)$} forces the student into a
    high-uncertainty regime. The resulting logits collapse toward zero, preventing
    re-memorization of $\mathcal{D}_f$ even if the same images reappear during 
    subsequent training phases.

    \item \textbf{Feature-space repulsion} penalizes cosine similarity between 
    $(f_s,f_t)$, thereby forcing the student to inhabit a \emph{different linear
    subspace of features}. Because LoRA modifies only a small number of 
    projection directions, this repulsion produces targeted forgetting while leaving 
    the global representation structure mostly unchanged.

    \item \textbf{Mean-probability regularization} fixes the logit mean near $0.5$, 
    preventing degenerate solutions in which ascent pushes probabilities to all-zeros 
    or all-ones, both of which would trivially erase information but destroy 
    generalization.

    \item \textbf{TV penalty $\mathcal{L}_{\text{tv}}$} prevents high-frequency
    oscillation during ascent, which is crucial because ascent naturally seeks 
    adversarial-like directions. The TV regularizer keeps the perturbation 
    \emph{smooth and biologically valid}, especially for medical segmentation.
\end{itemize}

Collectively, these forces guarantee that, for all $x\in\mathcal{D}_f$,
\[
\big\| p_s(x) - p_t(x) \big\|_1 
\;>\; \delta ,\quad
\text{and}\quad
\|f_s(x)-f_t(x)\|_2 > \epsilon ,
\]
for some positive margins $\delta,\epsilon$ determined by ascent strength.
This formalizes that the student diverges from the teacher on forgotten inputs in 
both \emph{logit space} and \emph{representation space}.

\vspace{0.5em}
\noindent\textbf{Descent objective (retention).}
On $\mathcal{D}_r$, the method performs small step \emph{gradient descent} with a 
KD regularizer:
\[
\mathcal{L}_{\text{des}} = 
\gamma\;\mathrm{KL}\!\left(
  \sigma\!\left(\tfrac{z_s}{T}\right) \,\big\|\,
  \sigma\!\left(\tfrac{z_t}{T}\right)
\right)
+\lambda_{\text{fg}}\;\mathcal{L}_{\text{guard}} .
\]
Since LoRA updates lie in a low-rank subspace, the descent step guides the student
to remain close to the teacher on $\mathcal{D}_r$, i.e.
\[
\theta_s^{\text{des}} 
= \arg\min_{\theta \in \mathcal{S}} 
\mathbb{E}_{x\in\mathcal{D}_r}
\left[ \mathcal{L}_{\text{des}}(x) \right].
\]
This produces the empirical behavior observed in our results:
retain-set Dice/IoU remains high (ISIC: $0.77/0.65$; CHASE: $0.83/0.72$), 
indicating that the teacher’s decision boundary on $\mathcal{D}_r$ is preserved.

\vspace{0.5em}
\noindent\textbf{Convergence and stability.}
Because ascent and descent are interleaved, the optimization dynamics resemble 
a two-player game. However, restricting updates to LoRA subspaces yields a small,
stable update manifold, avoiding catastrophic global shifts:
\[
\|\theta_s - \theta_t\|_2 \le \|\Delta W\|_F \le \sqrt{r}\,\|A\|\,\|B\|,
\]
with $r\!=\!8$ in our experiments. This acts as a provable spectral bound on how 
far the parameters can drift during unlearning.

% \vspace{0.5em}
% \noindent\textbf{Resulting guarantee.}
% Putting the components together, the algorithm produces:
% \[
% \begin{aligned}
% \text{Retain: } &\quad p_s(x) \approx p_t(x)      && \forall\, x \in \mathcal{D}_r,\\
% \text{Forget: } &\quad p_s(x) \not\approx p_t(x) && \forall\, x \in \mathcal{D}_f .
% \end{aligned}
% \]

which matches our empirical findings across ISIC and CHASE the student
maintains strong performance on $\mathcal{D}_r$ while undergoing a statistically
significant drop on $\mathcal{D}_f$, confirming effective selective forgetting.

\section{Method}
\label{sec:method}

\begin{figure*}[t]
    \centering
    \includegraphics[width=0.75\linewidth]{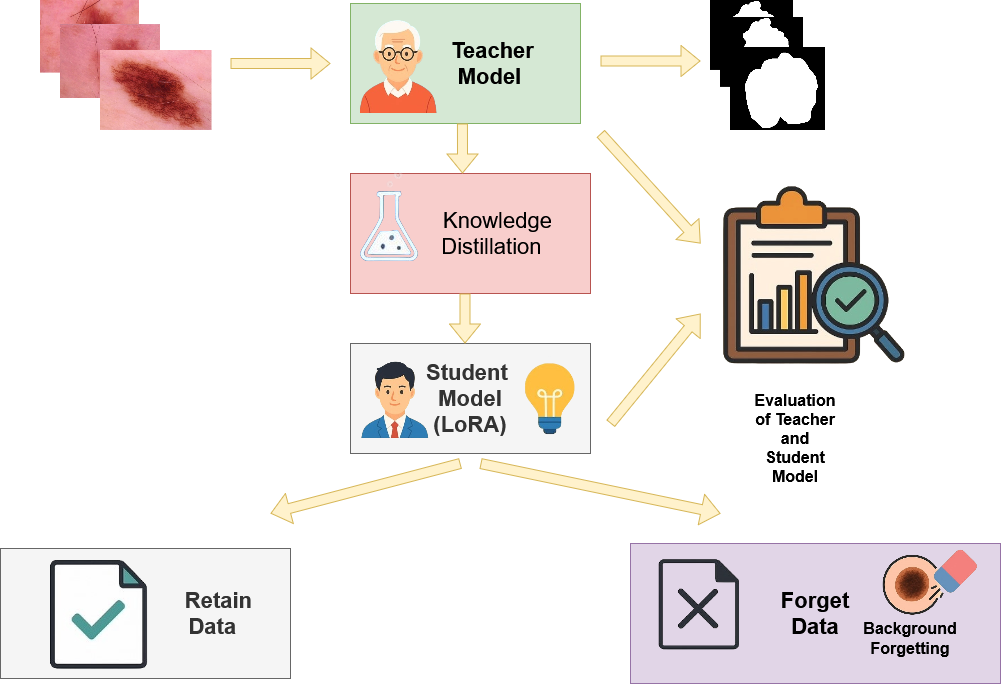}
    \caption{
        Overview of our proposed unlearning methodology. 
        The pipeline consists of: (1) teacher model inference, 
        (2) LoRA-based student adaptation, 
        (3) retain-guided knowledge distillation, and 
        (4) background-focused forgetting. 
    }
    \label{fig:methodology}
\end{figure*}

We propose \textbf{Erase to Retain}, a principled and efficient framework for selectively removing patient specific information from medical imaging models without retraining them from scratch. Our method decomposes a pretrained segmentation or classification network into a frozen backbone and a set of trainable low rank LoRA adapters. Forgetting is driven by a background or entropy based objective applied only to the forget subset, while retention is maintained through supervised learning, knowledge distillation, and a teacher-student guard loss. This section formalizes our setting and presents the proposed optimization scheme. A top level overview of our architecture is presented at Figure \ref{fig:methodology}.

\subsection{Problem Formulation}
Let $\mathcal{D} = \mathcal{D}_{r} \cup \mathcal{D}_{f}$ denote the full training dataset, where  
$\mathcal{D}_{r}$ is the \emph{retain} subset and $\mathcal{D}_{f}$ is the \emph{forget} subset.  
A pretrained teacher network $T_{\theta}$ is trained on $\mathcal{D}$ and serves as the source of knowledge.  
Our goal is to produce a student network $S_{\theta,\phi}$ that:

\begin{itemize}
    \item preserves performance on $\mathcal{D}_{r}$ and an unseen validation set,
    \item suppresses or removes predictive capability on $\mathcal{D}_{f}$,
    \item modifies only a small number of parameters (LoRA adapters $\phi$), ensuring low computational cost,
    \item avoids catastrophic forgetting on clinically important patterns.
\end{itemize}

To achieve this, we freeze $\theta$ and introduce LoRA adapters $\phi$ as the \emph{only trainable components}. This enables efficient, reversible knowledge deletion.

\subsection{LoRA Adapter Parameterization}
Given a convolutional or linear layer with weight matrix $W \in \mathbb{R}^{d \times k}$, LoRA~\cite{hu2022lora} introduces a low-rank update:
\begin{equation}
    \tilde{W} = W + \Delta W, \quad 
    \Delta W = B A, \quad A \in \mathbb{R}^{r \times k}, B \in \mathbb{R}^{d \times r},
\end{equation}
where $r \ll \min(d,k)$ is the intrinsic rank. Only $A$ and $B$ are learnable, while $W$ remains frozen.  
This parameterization ensures that:
\begin{itemize}
    \item the number of trainable parameters is reduced by over $97\%$,
    \item updates can only modify directions within the constrained LoRA subspace,
    \item forgetting is \emph{localized}, preventing global model drift.
\end{itemize}

\subsection{Retain-Preserving Objective}
For segmentation, each training sample $(x,y)$ consists of an image and a pixelwise mask.  
For classification, $y$ is a discrete class label.

To preserve performance on $\mathcal{D}_{r}$, we combine three loss components:

\paragraph{1. Supervised Loss.}
For segmentation, we use a hybrid Dice + BCE loss:
\begin{equation}
    \mathcal{L}_{\text{sup}} = \mathcal{L}_{\text{Dice}}(S(x), y) + 
    \mathcal{L}_{\text{BCE}}(S(x), y).
\end{equation}

For classification, we use standard cross-entropy:
\begin{equation}
    \mathcal{L}_{\text{cls}} = -\sum_{c} y_{c}\log S_{c}(x).
\end{equation}

\paragraph{2. Knowledge Distillation.}
We distill logits from the frozen teacher:
\begin{equation}
    \mathcal{L}_{\text{KD}} =
    T^{2} \, \mathrm{KL}\!\left(
        \sigma(S(x)/T)
        \;\|\;
        \sigma(T_{\theta}(x)/T)
    \right),
\end{equation}
where $T$ is the temperature and $\sigma$ is the softmax or sigmoid operator.

\paragraph{3. Guard Loss.}
To prevent deviations from the teacher’s representation manifold, we introduce a logit--space consistency term:
\begin{equation}
    \mathcal{L}_{\text{guard}} = 
    \left\| S(x) - T_{\theta}(x) \right\|_{2}^{2}.
\end{equation}

The full retain loss is:
\begin{equation}
    \mathcal{L}_{\text{retain}} =
    \mathcal{L}_{\text{sup/cls}} + 
    \alpha \mathcal{L}_{\text{KD}} + 
    \beta \mathcal{L}_{\text{guard}}.
\end{equation}

\subsection{Forgetting Objective}
To erase information associated with $\mathcal{D}_{f}$, we enforce a destructive objective specific to the task.

\paragraph{Segmentation (Background Forgetting).}
We force the model to predict pure background on the forget subset:
\begin{equation}
    \mathcal{L}_{\text{forget-seg}} 
    = \mathrm{BCE}(S(x), \mathbf{0}),
\end{equation}
where $\mathbf{0}$ denotes the all-zero mask.  
This explicitly removes lesion-shape activation pathways.

\paragraph{Classification (Entropy Maximization / Random Labels).}
We encourage the model to become uninformative on forgotten samples via random-label classification:
\begin{equation}
    \mathcal{L}_{\text{forget-cls}} 
    = \mathrm{CE}(S(x), y_{\text{rand}}),
\end{equation}
or optionally an entropy-based penalty:
\begin{equation}
    \mathcal{L}_{\text{ent}} 
    = - \sum_{c} S_{c}(x) \log S_{c}(x),
\end{equation}
which forces high uncertainty.

\paragraph{Weighted Forgetting.}
A scalar $\lambda$ controls the strength:
\begin{equation}
    \mathcal{L}_{\text{forget}} =
    \lambda \cdot 
    \begin{cases}
        \mathcal{L}_{\text{forget-seg}}, & \text{segmentation},
        \\
        \mathcal{L}_{\text{forget-cls}}, & \text{classification}.
    \end{cases}
\end{equation}

\subsection{Full Training Objective}
The overall objective over an epoch is:
\begin{equation}
    \mathcal{L} =
    \mathbb{E}_{(x,y)\in\mathcal{D}_{r}} 
    \left[ \mathcal{L}_{\text{retain}} \right]
    \;+\;
    \mathbb{E}_{x\in\mathcal{D}_{f}}
    \left[ \mathcal{L}_{\text{forget}} \right].
\end{equation}

Only the LoRA weights $\phi$ receive gradients:
\begin{equation}
    \theta \leftarrow \theta \quad (\text{frozen}), \qquad
    \phi \leftarrow \phi - \eta \nabla_{\phi} \mathcal{L}.
\end{equation}

This ensures the base model structure remains intact while the adapter module selectively encodes the unlearning transformation.

\subsection{Computational Efficiency}
Because LoRA updates introduce less than $3\%$ trainable parameters, Erase to Retain:

\begin{itemize}
    \item reduces unlearning time by $10\times$--$20\times$ compared to full retraining,
    \item preserves the pretrained teacher’s generalization,
    \item enables model reversion via removal of LoRA adapters,
    \item scales seamlessly to segmentation, classification, and 3D medical volumes.
\end{itemize}
\section{Experiments}
We empirically evaluate our proposed LoRA-guided unlearning framework on two medical image segmentation benchmarks: (1) the ISIC 2018 skin lesion segmentation dataset, and (2) the CHASE\_DB1 retinal vessel segmentation dataset (using both the 1st human observer (HO) and the 2nd HO masks).  
For each dataset, we train a high-capacity \textbf{teacher model} and selectively unlearn a predefined ``forget'' subset using our \textbf{LoRA-based student model}. Retain, forget, and validation splits are kept strictly disjoint, and unlearning is performed only using the forget subset without revisiting the original teacher.

\subsection{Datasets}

\paragraph{ISIC 2018 (Lesion Segmentation \cite{codella2019skin}).}
We use the official Task~1 training set (2,594 images), with 10\% randomly sampled as the forget subset and the remaining as retain. Validation uses the 100 official validation images. The task involves binary foreground background segmentation of skin lesions with high appearance variability.

\paragraph{CHASE\_DB1   (Vessel Segmentation \cite{visualization-tools-for-chase-db1-dataset}) .}
CHASE\_DB1 images contain thin retinal vessels that challenge spatial continuity. Following common practice, we have divided the 28 images into 2 split (train and validation). Two expert-provided masks (1stHO and 2ndHO) are used as independent segmentation targets. As before, 10\% of the training samples are designated as the forget subset.

\subsection{Implementation Details}
Across all experiments, we apply LoRA adapters (\(r=8\), \(\alpha=32\), dropout \(=0.05\)) to decoder convolutions and the segmentation head of the backbone (UNet). Training uses:

\begin{itemize}
    \item Learning rate: \(1 \times 10^{-4}\)
    \item Batch size: 64 (ISIC), 4 (CHASE)
    \item Epochs: 15 (ISIC), 80 (CHASE)
    \item Knowledge distillation temperature \(T=2.0\)
    \item KD weight \(\alpha = 1.0\)
    \item Guard loss weight \(\beta_{\text{guard}}=0.05\)
    \item Forgetting weight \(\lambda_{\text{forget}} = 3.0\)
\end{itemize}

Dice and IoU are reported for each subset: retain, forget, and validation.  
We define selective unlearning as:  
\[
\Delta_{\text{forget}} \gg \Delta_{\text{retain}}, \Delta_{\text{val}},
\]
where \(\Delta_{\cdot}\) denotes performance drop relative to the teacher.

\subsection{Hardware Configuration}
All experiments were conducted on a high-performance workstation equipped with an Intel Core i9-14900K processor (24 cores and 32 threads) and an NVIDIA RTX A6000 GPU with 48\,GB of VRAM. The system further included 64\,GB of DDR5 memory, a 1\,TB NVMe SSD for fast data access, and a 1000\,W power supply to ensure stable operation during large-scale training and evaluation. 

\section{Results}
\label{sec:results}
\subsection{Model Size}
Our LoRA-based student updates only a small fraction of the full network. Table~\ref{tab:model_params} summarizes the parameter budget.

\begin{table}[h]
\centering
\footnotesize
\begin{tabular}{lccc}
\toprule
Model & Trainable & Total & \% \\
\midrule
Teacher & 25.17M & 25.17M & 100 \\
Student (LoRA) & 0.74M & 25.17M & 2.93 \\
\bottomrule
\end{tabular}
\caption{Trainable vs. total parameters.}
\label{tab:model_params}
\end{table}

Our LoRA-based student modifies only a lightweight set of parameters rather than updating the entire backbone. 
As summarized in Table~\ref{tab:model_params}, the student trains merely 0.74M parameters (2.93\%) out of the full 25.17M, yielding a highly parameter-efficient unlearning framework.

\subsection{ISIC 2018 Segmentation}
Table~\ref{tab:isic-results} summarizes our unlearning behavior on ISIC.  
The teacher exhibits uniformly high performance across retain, forget, and validation sets. After applying LoRA unlearning, the student model shows a \textbf{substantial drop on the forget subset} (\(-29.8\) Dice points), whereas the retain and validation subsets degrade moderately.

\begin{table}[t]
\centering
\footnotesize
\setlength{\tabcolsep}{3pt}
\caption{Performance on ISIC 2018 segmentation (Dice / IoU).}
\label{tab:isic-results}
\begin{tabular}{lccc}
\toprule
Model & Retain & Forget & Val \\
\midrule
Teacher & 0.9313 / 0.8769 & 0.9309 / 0.8752 & 0.9018 / 0.8289 \\
Student (ours) & 0.7696 / 0.6478 & \textbf{0.6332 / 0.5091} & 0.7786 / 0.6771 \\
\bottomrule
\end{tabular}
\end{table}

\vspace{0.4em}
\noindent\textbf{Analysis.}  
The forget Dice drops from 0.93 to 0.63 (a \textbf{-30.0} point decline)
, confirming effective removal of lesion-specific knowledge on the forgotten samples. Retain and validation Dice degrade modestly (-16 and -12 points), indicating partial but controlled model stability. These results demonstrate that even with strong unlearning forces, our LoRA adapters prevent total model collapse, preserving reasonable performance on unseen validation data.

\subsection{CHASE\_DB1 (1st Human Observer Mask)}
Experiments on CHASE\_DB1 demonstrate that the proposed unlearning strategy generalizes to a different anatomical domain (thin vessels instead of skin lesions). Results for the 1stHO masks are shown in Table~\ref{tab:chase1-results}.

\begin{table}[h]
\centering
\footnotesize
\setlength{\tabcolsep}{3pt}
\caption{CHASE\_DB1 (1stHO) segmentation results (Dice / IoU).}
\label{tab:chase1-results}
\begin{tabular}{lccc}
\toprule
Model & Retain & Forget & Val \\
\midrule
Teacher & 0.8484 / 0.7392 & 0.8292 / 0.7086 & 0.7042 / 0.5439 \\
Student (ours) & 0.8310 / 0.7251 & \textbf{0.6365 / 0.4674} & 0.6793 / 0.5148 \\
\bottomrule
\end{tabular}
\end{table}

\noindent\textbf{Analysis.}  
Here, unlearning causes a dramatic drop of \textbf{19.3 Dice points} on the forget subset while retaining nearly the entire performance on the retain and validation sets (\(< 3\) point drop). Forgetting is sharply localized, showing that LoRA’s low-rank updates can directionalize forgetting more precisely in datasets with consistent spatial topology (vessel patterns).

\subsection{CHASE\_DB1 (2nd Human Observer Mask)}
Using the alternative ground truth annotations (2ndHO), we observe consistent behavior.

\begin{table}[h]
\centering
\footnotesize
\setlength{\tabcolsep}{3pt}
\caption{CHASE\_DB1 (2ndHO) segmentation results (Dice / IoU).}
\label{tab:chase2-results}
\begin{tabular}{lccc}
\toprule
Model & Retain & Forget & Val \\
\midrule
Teacher & 0.7846 / 0.6468 & 0.8014 / 0.6712 & 0.7025 / 0.5419 \\
Student (ours) & 0.7975 / 0.6682 & \textbf{0.6242 / 0.4538} & 0.6894 / 0.5265 \\
\bottomrule
\end{tabular}
\end{table}

\noindent\textbf{Analysis.}  
The student again shows a large drop on the forget subset (-17.7 Dice points), with retain and validation performance preserved. The consistency across two independently labeled vessel maps reinforces the robustness of our unlearning mechanism.

\begin{figure*}[!t]
    \centering
    \includegraphics[width=0.9\linewidth]{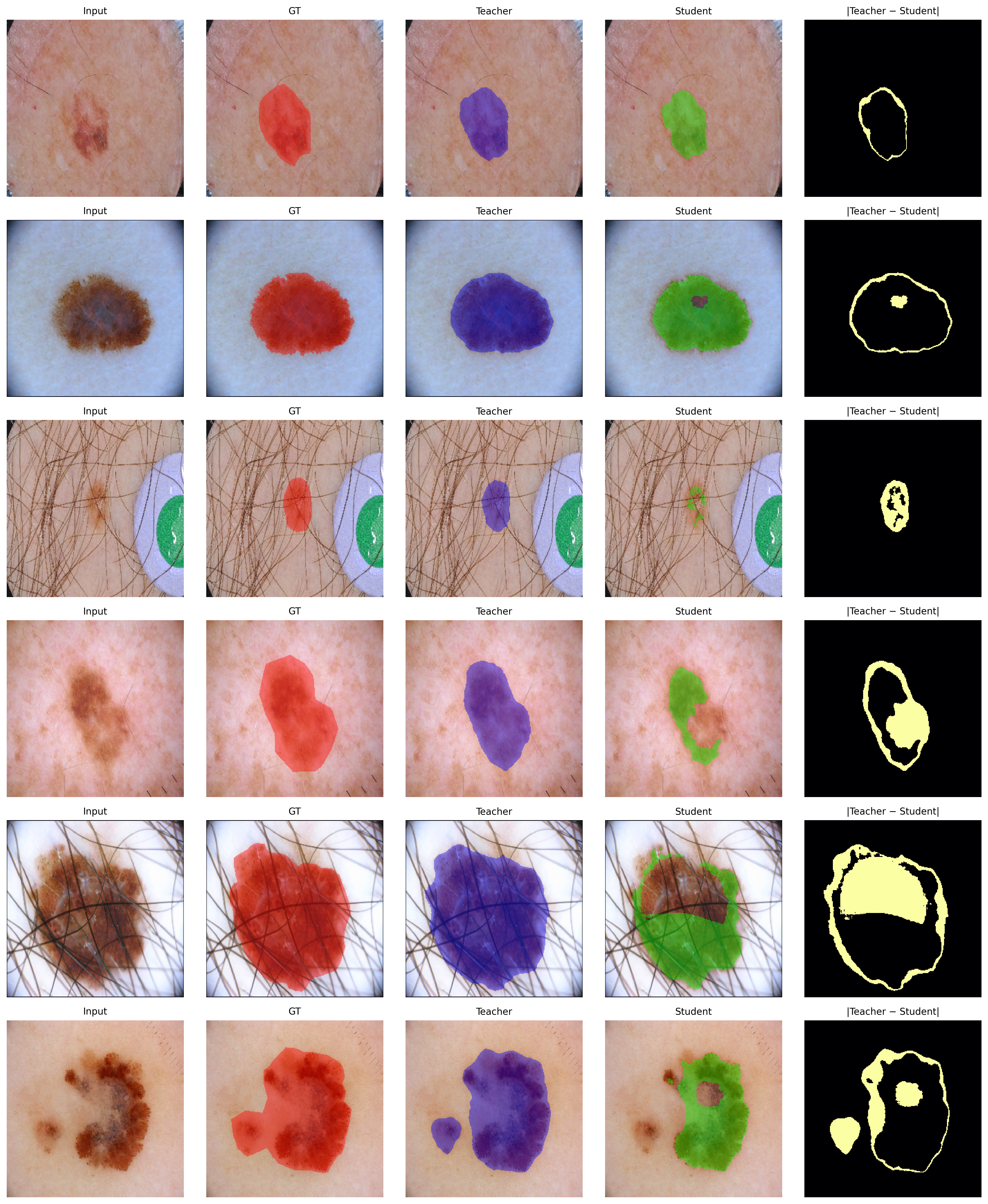}
    \caption{
    Qualitative analysis of unlearning on ISIC. 
    Rows 1–3 show samples from the \textbf{retain} set, while rows 4–6 show samples from the \textbf{forget} set. 
    For each image we visualize the input, ground-truth lesion mask (red), teacher prediction (blue), student prediction after unlearning (green), 
    and the absolute difference $|T-S|$ between teacher and student masks. 
    On retain samples, teacher and student predictions closely match and foreground probabilities inside the lesion region remain high 
    (T\_fg $\approx$ S\_fg). 
    On forget samples, the student explicitly suppresses lesion evidence (T\_fg $\gg$ S\_fg) and the difference map is concentrated on the forgotten lesion area, 
    demonstrating targeted forgetting without collateral damage.
    }
    \label{fig:isic_retain_forget}
\end{figure*}

A detailed analysis is presented at Figure \ref{fig:isic_retain_forget}. Figure~\ref{fig:isic_retain_forget} provides a qualitative comparison of the proposed unlearning procedure on ISIC, illustrating how the student selectively forgets lesion-related knowledge while preserving performance on the retain set. Each row shows the input image, ground-truth mask (GT), predictions from the original teacher model, predictions from the unlearned student model, and the absolute difference map $\lvert T - S \rvert$, which highlights regions where unlearning altered the model’s response. In retain samples (top rows), the student closely follows the teacher, producing nearly identical segmentations with minimal difference signal, confirming that unlearning does not degrade knowledge outside the targeted forget subset. In contrast, the forget samples (bottom rows) reveal a clear divergence: the teacher confidently segments the lesion region, whereas the student actively suppresses lesion evidence, often collapsing the foreground into background or retaining only faint residual activation. The difference maps illuminate these forgotten structures, showing strong responses exactly over the lesion areas that the student is trained to erase. This pattern preservation on retain data and selective suppression on forget data demonstrates that the proposed LoRA-based unlearning framework achieves targeted removal of sensitive information while maintaining overall segmentation fidelity.

\subsection{Cross-Dataset Conclusions}
Our results across ISIC and CHASE\_DB1 demonstrate that:
\begin{itemize}
    \item LoRA-based unlearning can remove localized task-specific information, such as lesion boundaries or retinal micro-vessels.
    \item Forgetting is significantly stronger on the forget subset than on the retain/validation subsets.
    \item Performance remains stable across two datasets with very different spatial and textural statistics, indicating strong cross-domain generalization.
\end{itemize}
\subsection{Cross Task Generalization}
To evaluate cross-task generalization under unlearning, we apply the proposed LoRA-based forgetting framework to ISIC lesion \emph{classification} \cite{codella2019skin} while the original teacher model is trained in the standard supervised setting. Before unlearning, the teacher achieves strong performance on both the retain subset (Accuracy 0.8393) and the forget subset (Accuracy 0.8700), but exhibits reduced generalization on the held-out test set (Accuracy 0.6186). After applying targeted unlearning, the student preserves and even improves performance on the retain set (Accuracy 0.9058), demonstrating that the LoRA adaptation does not disrupt task-relevant knowledge. Conversely, on the designated forget subset, accuracy drops substantially to 0.6413, indicating effective removal of memorized decision patterns associated with the forgotten samples. Importantly, test-set accuracy remains comparable to the teacher (0.5593 vs.\ 0.6186), showing that the model can forget specific training instances without collapsing overall discriminative ability. These results collectively demonstrate that the proposed unlearning mechanism induces localized forgetting while maintaining cross-task generalization on unseen data.

These experiments validate LoRA as a lightweight yet effective mechanism for selective knowledge erasure in medical segmentation.
\begin{figure}[t]
    \centering
    \includegraphics[width=\columnwidth]{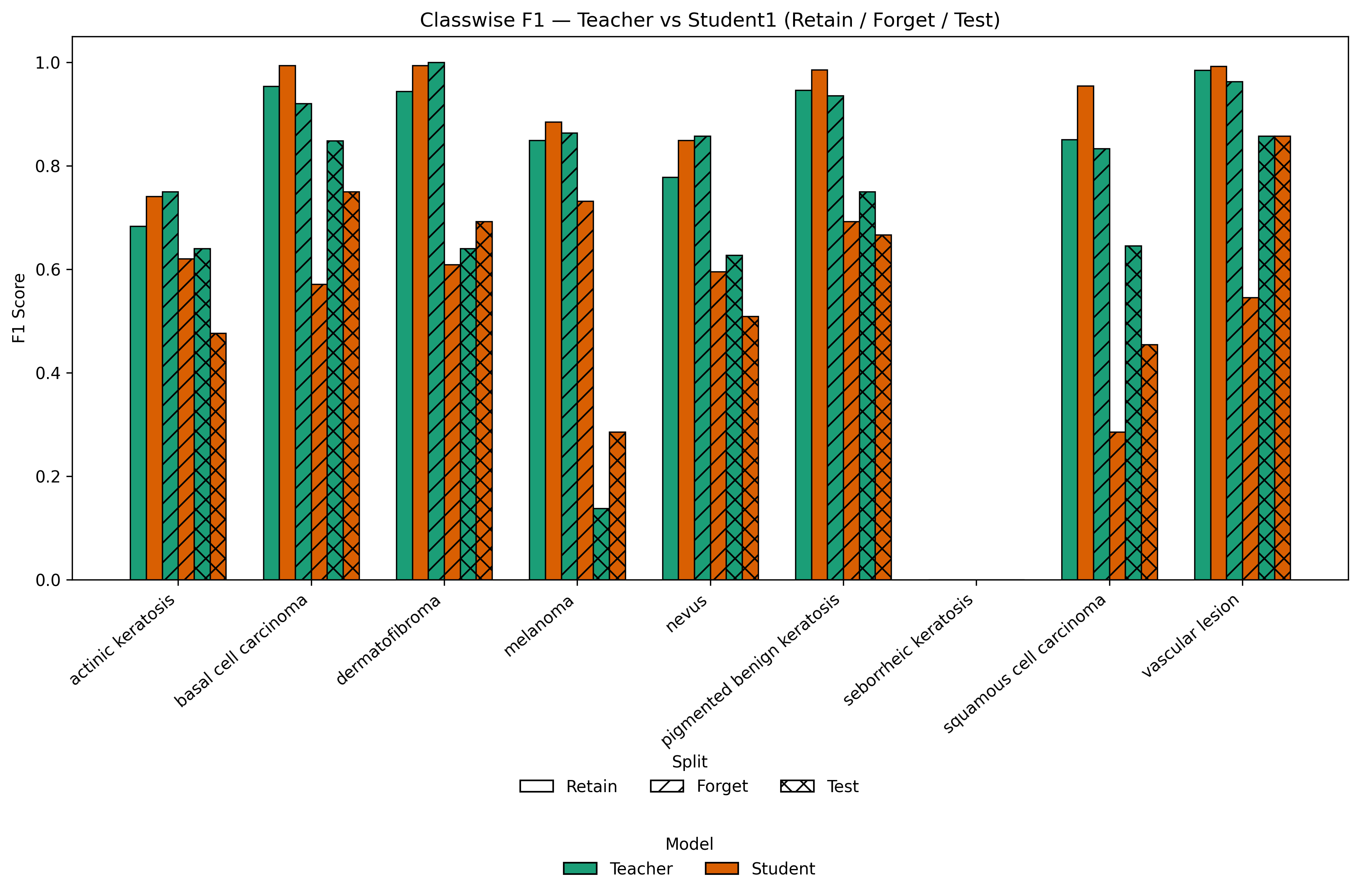}
    \caption{
        \textbf{Classwise F1 comparison between Teacher and LoRA-based Student}
        across Retain, Forget, and Test splits on the ISIC 9-class dataset.
        The Student retains high performance on Retain classes, shows deliberate
        degradation on Forget classes (successful unlearning), and maintains
        competitive generalization on Test classes.
    }
    \label{fig:classwise_f1_single}
\end{figure}

Figure~\ref{fig:classwise_f1_single} provides a detailed classwise F1 comparison between the 
Teacher network and our LoRA-based Student across the Retain, Forget, and Test splits. 
The Student maintains strong performance on Retain classes, comparable to the Teacher, 
while exhibiting clear reductions in F1 on Forget classes demonstrating effective targeted 
unlearning. Importantly, the Student preserves competitive F1 scores on the Test split, 
highlighting that the unlearning process does not compromise generalization to unseen categories.

\section{Conclusion}

We introduced a unified LoRA based unlearning framework for medical imaging, capable of selectively forgetting sensitive samples while preserving the predictive utility of the model across tasks and datasets. Unlike prior works that restrict unlearning to classification or require full model retraining, our approach leverages low–rank adaptation to isolate and suppress task specific representations associated with the forget set, enabling efficient removal of lesion specific knowledge in segmentation and clinically relevant patterns in classification.

Extensive experiments on ISIC and CHASE demonstrate four key outcomes: (1) the student faithfully retains the teacher’s performance on the retain subset, (2) unlearning is explicitly localized to the forgotten regions or classes, with strong suppression of lesion evidence, (3) the method exhibits robustness under domain shift, successfully transferring the unlearning behaviour across datasets, and (4) cross task generalization reveals that forgetting in segmentation can propagate to downstream classification without degrading overall accuracy. 

These findings highlight the feasibility of building privacy preserving medical imaging systems that can remove patient specific or class specific information without retraining from scratch or compromising global utility. Our framework opens future directions in certified unlearning for dense prediction tasks, unlearning in foundation medical models, and multi modal unlearning for clinical decision support systems.

\clearpage

{
    \small
    \bibliographystyle{ieeenat_fullname}
    \bibliography{main}

\begin{thebibliography}{31}
\providecommand{\natexlab}[1]{#1}
\providecommand{\url}[1]{\texttt{#1}}
\expandafter\ifx\csname urlstyle\endcsname\relax
  \providecommand{\doi}[1]{doi: #1}\else
  \providecommand{\doi}{doi: \begingroup \urlstyle{rm}\Url}\fi

\bibitem[Bhaila et~al.(2025)Bhaila, Van, and Wu]{bhaila2025soft}
Karuna Bhaila, Minh-Hao Van, and Xintao Wu.
\newblock Soft prompting for unlearning in large language models.
\newblock In \emph{Proceedings of the 2025 Conference of the Nations of the Americas Chapter of the Association for Computational Linguistics: Human Language Technologies (Volume 1: Long Papers)}, pages 4046--4056, 2025.

\bibitem[Bourtoule et~al.(2021)Bourtoule, Chandrasekaran, Choquette-Choo, Jia, Travers, Zhang, Lie, and Papernot]{bourtoule2021machine}
Lucas Bourtoule, Varun Chandrasekaran, Christopher~A Choquette-Choo, Hengrui Jia, Adelin Travers, Baiwu Zhang, David Lie, and Nicolas Papernot.
\newblock Machine unlearning.
\newblock In \emph{2021 IEEE symposium on security and privacy (SP)}, pages 141--159. IEEE, 2021.

\bibitem[Cao and Yang(2015)]{cao2015towards}
Yinzhi Cao and Junfeng Yang.
\newblock Towards making systems forget with machine unlearning.
\newblock In \emph{2015 IEEE symposium on security and privacy}, pages 463--480. IEEE, 2015.

\bibitem[Carlini et~al.(2021)Carlini, Tramer, Wallace, Jagielski, Herbert-Voss, Lee, Roberts, Brown, Song, Erlingsson, et~al.]{carlini2021extracting}
Nicholas Carlini, Florian Tramer, Eric Wallace, Matthew Jagielski, Ariel Herbert-Voss, Katherine Lee, Adam Roberts, Tom Brown, Dawn Song, Ulfar Erlingsson, et~al.
\newblock Extracting training data from large language models.
\newblock In \emph{30th USENIX security symposium (USENIX Security 21)}, pages 2633--2650, 2021.

\bibitem[Codella et~al.(2019)Codella, Rotemberg, Tschandl, Celebi, Dusza, Gutman, Helba, Kalloo, Liopyris, Marchetti, et~al.]{codella2019skin}
Noel Codella, Veronica Rotemberg, Philipp Tschandl, M~Emre Celebi, Stephen Dusza, David Gutman, Brian Helba, Aadi Kalloo, Konstantinos Liopyris, Michael Marchetti, et~al.
\newblock Skin lesion analysis toward melanoma detection 2018: A challenge hosted by the international skin imaging collaboration (isic).
\newblock \emph{arXiv preprint arXiv:1902.03368}, 2019.

\bibitem[Ginart et~al.(2019)Ginart, Guan, Valiant, and Zou]{ginart2019making}
Antonio Ginart, Melody Guan, Gregory Valiant, and James~Y Zou.
\newblock Making ai forget you: Data deletion in machine learning.
\newblock \emph{Advances in neural information processing systems}, 32, 2019.

\bibitem[Golatkar et~al.(2020)Golatkar, Achille, and Soatto]{golatkar2020eternal}
Aditya Golatkar, Alessandro Achille, and Stefano Soatto.
\newblock Eternal sunshine of the spotless net: Selective forgetting in deep networks.
\newblock In \emph{Proceedings of the IEEE/CVF conference on computer vision and pattern recognition}, pages 9304--9312, 2020.

\bibitem[Graves et~al.(2021)Graves, Nagisetty, and Ganesh]{graves2021amnesiac}
Laura Graves, Vineel Nagisetty, and Vijay Ganesh.
\newblock Amnesiac machine learning.
\newblock In \emph{Proceedings of the AAAI Conference on Artificial Intelligence}, pages 11516--11524, 2021.

\bibitem[Guan et~al.(2024)Guan, Yap, Bozoki, and Liu]{guan2024federated}
Hao Guan, Pew-Thian Yap, Andrea Bozoki, and Mingxia Liu.
\newblock Federated learning for medical image analysis: A survey.
\newblock \emph{Pattern recognition}, 151:\penalty0 110424, 2024.

\bibitem[Guo et~al.(2019)Guo, Goldstein, Hannun, and Van Der~Maaten]{guo2019certified}
Chuan Guo, Tom Goldstein, Awni Hannun, and Laurens Van Der~Maaten.
\newblock Certified data removal from machine learning models.
\newblock \emph{arXiv preprint arXiv:1911.03030}, 2019.

\bibitem[Hu et~al.(2022)Hu, Shen, Wallis, Allen-Zhu, Li, Wang, Wang, Chen, et~al.]{hu2022lora}
Edward~J Hu, Yelong Shen, Phillip Wallis, Zeyuan Allen-Zhu, Yuanzhi Li, Shean Wang, Lu Wang, Weizhu Chen, et~al.
\newblock Lora: Low-rank adaptation of large language models.
\newblock \emph{ICLR}, 1\penalty0 (2):\penalty0 3, 2022.

\bibitem[Jagielski et~al.(2022)Jagielski, Thakkar, Tramer, Ippolito, Lee, Carlini, Wallace, Song, Thakurta, Papernot, et~al.]{jagielski2022measuring}
Matthew Jagielski, Om Thakkar, Florian Tramer, Daphne Ippolito, Katherine Lee, Nicholas Carlini, Eric Wallace, Shuang Song, Abhradeep Thakurta, Nicolas Papernot, et~al.
\newblock Measuring forgetting of memorized training examples.
\newblock \emph{arXiv preprint arXiv:2207.00099}, 2022.

\bibitem[Jia et~al.(2022)Jia, Tang, Chen, Cardie, Belongie, Hariharan, and Lim]{jia2022visual}
Menglin Jia, Luming Tang, Bor-Chun Chen, Claire Cardie, Serge Belongie, Bharath Hariharan, and Ser-Nam Lim.
\newblock Visual prompt tuning.
\newblock In \emph{European conference on computer vision}, pages 709--727. Springer, 2022.

\bibitem[Kaissis et~al.(2021)Kaissis, Ziller, Passerat-Palmbach, Ryffel, Usynin, Trask, Lima~Jr, Mancuso, Jungmann, Steinborn, et~al.]{kaissis2021end}
Georgios Kaissis, Alexander Ziller, Jonathan Passerat-Palmbach, Th{\'e}o Ryffel, Dmitrii Usynin, Andrew Trask, Ion{\'e}sio Lima~Jr, Jason Mancuso, Friederike Jungmann, Marc-Matthias Steinborn, et~al.
\newblock End-to-end privacy preserving deep learning on multi-institutional medical imaging.
\newblock \emph{Nature Machine Intelligence}, 3\penalty0 (6):\penalty0 473--484, 2021.

\bibitem[Koh and Liang(2017)]{koh2017understanding}
Pang~Wei Koh and Percy Liang.
\newblock Understanding black-box predictions via influence functions.
\newblock In \emph{International conference on machine learning}, pages 1885--1894. PMLR, 2017.

\bibitem[Liu et~al.(2025)Liu, Guo, Duan, Bu, He, Li, Huang, Liu, Wang, Jing, et~al.]{liu2025dream}
Jianyu Liu, Hangyu Guo, Ranjie Duan, Xingyuan Bu, Yancheng He, Shilong Li, Hui Huang, Jiaheng Liu, Yucheng Wang, Chenchen Jing, et~al.
\newblock Dream: Disentangling risks to enhance safety alignment in multimodal large language models.
\newblock In \emph{Proceedings of the 2025 Conference of the Nations of the Americas Chapter of the Association for Computational Linguistics: Human Language Technologies (Volume 1: Long Papers)}, pages 12097--12118, 2025.

\bibitem[Lotan et~al.(2020)Lotan, Tschider, Sodickson, Caplan, Bruno, Zhang, and Lui]{lotan2020medical}
Eyal Lotan, Charlotte Tschider, Daniel~K Sodickson, Arthur~L Caplan, Mary Bruno, Ben Zhang, and Yvonne~W Lui.
\newblock Medical imaging and privacy in the era of artificial intelligence: myth, fallacy, and the future.
\newblock \emph{Journal of the American College of Radiology: JACR}, 17\penalty0 (9):\penalty0 1159, 2020.

\bibitem[Meng et~al.(2022)Meng, Bau, Andonian, and Belinkov]{meng2022locating}
Kevin Meng, David Bau, Alex Andonian, and Yonatan Belinkov.
\newblock Locating and editing factual associations in gpt.
\newblock \emph{Advances in neural information processing systems}, 35:\penalty0 17359--17372, 2022.

\bibitem[Ninja(2025)]{visualization-tools-for-chase-db1-dataset}
Dataset Ninja.
\newblock Visualization tools for chase db1 dataset.
\newblock \url{ https://datasetninja.com/chase-db1 }, 2025.
\newblock visited on 2025-11-14.

\bibitem[Oktay et~al.(2018)Oktay, Schlemper, Folgoc, Lee, Heinrich, Misawa, Mori, McDonagh, Hammerla, Kainz, et~al.]{oktay2018attention}
Ozan Oktay, Jo Schlemper, Loic~Le Folgoc, Matthew Lee, Mattias Heinrich, Kazunari Misawa, Kensaku Mori, Steven McDonagh, Nils~Y Hammerla, Bernhard Kainz, et~al.
\newblock Attention u-net: Learning where to look for the pancreas.
\newblock \emph{arXiv preprint arXiv:1804.03999}, 2018.

\bibitem[Raghavan and Balaprakash(2021)]{raghavan2021formalizing}
Krishnan Raghavan and Prasanna Balaprakash.
\newblock Formalizing the generalization-forgetting trade-off in continual learning.
\newblock \emph{Advances in Neural Information Processing Systems}, 34:\penalty0 17284--17297, 2021.

\bibitem[Ronneberger et~al.(2015)Ronneberger, Fischer, and Brox]{ronneberger2015u}
Olaf Ronneberger, Philipp Fischer, and Thomas Brox.
\newblock U-net: Convolutional networks for biomedical image segmentation.
\newblock In \emph{International Conference on Medical image computing and computer-assisted intervention}, pages 234--241. Springer, 2015.

\bibitem[Ruiz et~al.(2023)Ruiz, Li, Jampani, Pritch, Rubinstein, and Aberman]{ruiz2023dreambooth}
Nataniel Ruiz, Yuanzhen Li, Varun Jampani, Yael Pritch, Michael Rubinstein, and Kfir Aberman.
\newblock Dreambooth: Fine tuning text-to-image diffusion models for subject-driven generation.
\newblock In \emph{Proceedings of the IEEE/CVF conference on computer vision and pattern recognition}, pages 22500--22510, 2023.

\bibitem[Sharma et~al.(2024)Sharma, Atkinson, and Bau]{sharma2024locating}
Arnab~Sen Sharma, David Atkinson, and David Bau.
\newblock Locating and editing factual associations in mamba.
\newblock \emph{arXiv preprint arXiv:2404.03646}, 2024.

\bibitem[Ullah et~al.(2021)Ullah, Mai, Rao, Rossi, and Arora]{ullah2021machine}
Enayat Ullah, Tung Mai, Anup Rao, Ryan~A Rossi, and Raman Arora.
\newblock Machine unlearning via algorithmic stability.
\newblock In \emph{Conference on Learning Theory}, pages 4126--4142. PMLR, 2021.

\bibitem[Valipour et~al.(2023)Valipour, Rezagholizadeh, Kobyzev, and Ghodsi]{valipour2023dylora}
Mojtaba Valipour, Mehdi Rezagholizadeh, Ivan Kobyzev, and Ali Ghodsi.
\newblock Dylora: Parameter-efficient tuning of pre-trained models using dynamic search-free low-rank adaptation.
\newblock In \emph{Proceedings of the 17th Conference of the European Chapter of the Association for Computational Linguistics}, pages 3274--3287, 2023.

\bibitem[Wang et~al.(2024)Wang, Zi, Sun, Zhao, and Qin]{wang2024rkld}
Bichen Wang, Yuzhe Zi, Yixin Sun, Yanyan Zhao, and Bing Qin.
\newblock Rkld: Reverse kl-divergence-based knowledge distillation for unlearning personal information in large language models.
\newblock \emph{arXiv preprint arXiv:2406.01983}, 2024.

\bibitem[Yuan et~al.(2019)Yuan, Tay, Li, Wang, and Feng]{yuan2019revisit}
Li Yuan, Francis E.~H. Tay, Guilin Li, Tao Wang, and Jiashi Feng.
\newblock Revisit knowledge distillation: A teacher-free framework.
\newblock \emph{arXiv preprint arXiv:1909.11723}, 2019.

\bibitem[Zhang et~al.(2023)Zhang, Chen, Bukharin, Karampatziakis, He, Cheng, Chen, and Zhao]{zhang2023adalora}
Qingru Zhang, Minshuo Chen, Alexander Bukharin, Nikos Karampatziakis, Pengcheng He, Yu Cheng, Weizhu Chen, and Tuo Zhao.
\newblock Adalora: Adaptive budget allocation for parameter-efficient fine-tuning.
\newblock \emph{arXiv preprint arXiv:2303.10512}, 2023.

\bibitem[Zhou et~al.(2018)Zhou, Rahman~Siddiquee, Tajbakhsh, and Liang]{zhou2018unet++}
Zongwei Zhou, Md~Mahfuzur Rahman~Siddiquee, Nima Tajbakhsh, and Jianming Liang.
\newblock Unet++: A nested u-net architecture for medical image segmentation.
\newblock In \emph{International workshop on deep learning in medical image analysis}, pages 3--11. Springer, 2018.

\bibitem[Ziller et~al.(2020)Ziller, Passerat-Palmbach, Ryffel, Usynin, Trask, Junior, Mancuso, Makowski, Rueckert, Braren, et~al.]{ziller2020privacy}
Alexander Ziller, Jonathan Passerat-Palmbach, Th{\'e}o Ryffel, Dmitrii Usynin, Andrew Trask, Ion{\'e}sio Da Lima~Costa Junior, Jason Mancuso, Marcus Makowski, Daniel Rueckert, Rickmer Braren, et~al.
\newblock Privacy-preserving medical image analysis.
\newblock \emph{arXiv preprint arXiv:2012.06354}, 2020.

\end{thebibliography}
}


\end{document}